\documentclass[conference]{IEEEtran}
\IEEEoverridecommandlockouts
\usepackage{epsfig,makeidx,color,comment,subfigure}
\usepackage{amsmath,amssymb,bbm}
\usepackage{algpseudocode}
\usepackage{cite,graphicx,lipsum}
\usepackage{enumerate}
\usepackage{amsfonts}
\usepackage{booktabs}
\usepackage{siunitx}
\usepackage{empheq}
\usepackage[table,xcdraw]{xcolor}
\usepackage[switch,pagewise]{lineno}

\usepackage{hyperref}
\hypersetup{draft}
\usepackage{algorithm}

\floatname{algorithm}{Pseudocode}
\def\rT{{\rm T}}

\def\uR{{\mathbb R}}

\DeclareMathOperator*{\argmin}{\arg\!\min}

\newtheorem{mytheorem}{\bf Theorem} % [section]
\newtheorem{mylemma}{\bf Lemma} % [section]
\newtheorem{myexample}{\it Example} % [section]

\def\be{ \begin{equation} }
\def\ee{ \end{equation} }
\def\bea{ \begin{eqnarray} }
\def\eea{ \end{eqnarray} }
\def\bt{{\bf t}}
\def\bx{{\bf x}}
\def\by{{\bf y}}

\def\bd{{\bf d}}
\def\bb{{\bf b}}

\def\ba{{\bf a}}

\def\bu{{\bf u}}

\def\bz{{\bf z}}

\def\bv{{\bf v}}
\def\bw{{\bf w}}

\def\b0{{\bf 0}}

\def\cF{{\cal F}}

\def\cU{{\cal U}}
\def\cV{{\cal V}}
\def\cW{{\cal W}}

\def\sSNR{{\sf SNR}}

\algblock{Input}{EndInput}
\algnotext{EndInput}
\algblock{Output}{EndOutput}
\algnotext{EndOutput}
\algblock{Initializations}{EndInitializations}
\algnotext{EndInitializations}
\newcommand{\Desc}[2]{\State \makebox[2em][l]{#1}#2}

\ifCLASSOPTIONonecolumn
  \newcommand{\figwidth}{0.50\columnwidth}
  
\else
  \newcommand{\figwidth}{0.88\columnwidth}
  
\fi
\usepackage{graphicx}
\usepackage{booktabs}
\usepackage{graphicx}
\usepackage[normalem]{ulem}
\begin{document}

%\title{TSP with Neighborhoods for Wireless Data Collection by a UAV} 
% \title{Efficient Data Collection using UAVs by Exploiting Non-Zero Communication Ranges in Solving a Generalized TSP}
\title{Energy-Efficient UAV-Assisted IoT Data Collection via TSP-Based Solution Space Reduction}

\author{Sivaram Krishnan, Mahyar Nemati,
Seng W. Loke, Jihong Park, and Jinho Choi \\
\thanks{The authors are with
the School of Information Technology,
Deakin University, Geelong, VIC 3220, Australia
(e-mail address for the corrsponding author: jinho.choi@deakin.edu.au). This work was supported by the Institute of Information \& communications Technology Planning \& Evaluation (IITP) grant funded by the Korea government (MSIT) (No. 2021-0-00794, Development of 3D Spatial Mobile Communication Technology).
}}
\maketitle
\begin{abstract}

This paper presents a wireless data collection framework that employs an unmanned aerial vehicle (UAV) to efficiently gather data from distributed IoT sensors deployed in a large area. Our approach takes into account the non-zero communication ranges of the sensors to optimize the flight path of the UAV, resulting in a variation of the Traveling Salesman Problem (TSP). We prove mathematically that the optimal waypoints for this TSP-variant problem are restricted to the boundaries of the sensor communication ranges, greatly reducing the solution space. Building on this finding, we develop a low-complexity UAV-assisted sensor data collection algorithm, and demonstrate its effectiveness in a selected use case where we minimize the total energy consumption of the UAV and sensors by jointly optimizing the UAV's travel distance and the sensors' communication ranges.
% By minimizing the travel distance of the UAV, our framework provides an effective solution for data collection in diverse applications, as demonstrated in a case study 
\end{abstract}

\ifCLASSOPTIONonecolumn
\baselineskip 26pt
\fi

\section{Introduction}

Wireless sensor networks (WSNs) are typically made up of inexpensive Internet of Things (IoT) sensor nodes  that have been spatially distributed in order to detect and collect environmental and physical data sets
for various applications 
% \cite{Potdar09} \cite{Di11}.
including smart farming, battlefield monitoring, smart healthcare, and disaster warning.
However, these IoT sensor nodes may not be able to form a network (i.e., WSN) if the inter-node distances are long and the communication range of nodes is limited. Then, unmanned aerial vehicles (UAVs) can be used as data mules in transferring data swiftly over long distances, opportunistically \cite{nemati2021modelling} \cite{Tran22}. 
Such a scenario would require a UAV to dynamically move towards IoT sensor nodes, collect data and transmit this data to the base station (BS) or other sensor nodes which are typically outside the coverage radius. The employment of UAVs for data collecting is also motivated by the rising adaptability of 5G networks and wide bandwidth availability 
\cite{Access5G}  \cite{Nemati22}, which means that UAVs may operate at various altitudes and download enormous amounts of data from sensor nodes.

In particular,
with the advancement of wireless capabilities in UAVs in recent years, they have been used as data mules for data collection from sensor nodes. UAVs can be deployed to collect data from each sensor node with an energy-efficient approach. To begin data collection, a UAV must move close enough to a sensor node and does not need to hover directly over the sensor node; as a result, the link distance between the UAV and the sensor node is greatly reduced. We borrow the term ``data mule" from \cite{shah2003data}, a mobile object that can be used for data collection. The UAV's capabilities are not limited to data collection; it can also be used for surveillance and other tasks.

\subsection{Traveling Salesman Problem for UAV Path Planning}
%(2) review of UAV - TSP

The Traveling Salesman Problem (TSP) can be considered for UAV path planning \cite{sugiharagupta2011} \cite{kim2018traveling}. 
%\cite{Jang17} \cite{Chen17} 
Since the TSP is a nondeterministic polynomial (NP)-hard problem, there are a number of heuristic methods
such as simulated annealing \cite{bookstaber1997simulated}, genetic algorithm \cite{razali2011genetic}, and so on.
%tabu-search \cite{knox1994tabu}, 
%genetic algorithm \cite{razali2011genetic}, memetic algorithm \cite{merz2001memetic}, and so on.
While UAV path planning can be seen as a TSP, it would be necessary 
to take into consideration the communication range of the sensor node and the UAV. Since a UAV can simply pass through any point specified by the sensor node's communication range (also known as executable area or neighborhood) to collect data from the sensor, 
the TSP with neighborhoods (TSPN), which was introduced by \cite{arkin1994approximation}, can be considered for 
UAV path planning. TSP with a varying neighborhood size is studied in \cite{de2005tsp}, \cite{yuan2017towards}, while TSP with circular neighborhoods is studied in \cite{nedjatia2020robot}.In \cite{cheng2015data}, the authors proposed a method for efficient data collection in WSNs using a resource-constrained mobile sink. The approach modeled the TSP for minimizing the distance that the mobile sink has to travel and utilized reduce methods to develop a computationally efficient approach for data collection.

\subsection{Main Contributions and Organization} % of the Paper}

In this paper, while a TSPN is considered to model data collection by a UAV as in \cite{yuan2017towards} \cite{nedjatia2020robot}, the main contributions are as follows:
\begin{itemize}
%\item A UAV path planning problem to collect data from ground sensor nodes is formulated as a TSPN by exploiting a non-zero communication range of nodes. This problem is also studied in \cite{nedjatia2020robot} with non-overlapping communication regions, while we consider the cases where communication regions can be overlapped for closely located nodes.
\item It is rigorously proved that there exists at least
one optimal collection point that lies on the \emph{boundary} of each communication region of a node or the intersection of overlapped communication region of closely located nodes (to the best of our knowledge, this proof was previously unavailable), which can allow us to significantly reduce the travel distance in  UAV path planning.
%\item Based on the fact that it is suficient for the UAV to only  approach the boundary of the communication region, low-complexity approaches to UAV path planning are proposed. 
\item Through a case study, an optimization problem is studied to decide an optimal communication range when minimizing a weighted energy consumption subject to key constraints.
\end{itemize}

% SENG to continue here

\section{System Model}

%\subsection{A UAV for Data Collection from Multiple Nodes}

Suppose that there are $N$ ground nodes with measurements or data to send, which are distributed over a certain (remote) area. Each ground node has a limited energy source and its communication range is limited, meaning that each node is unable to transmit its signals to an access point connected to a backbone network.
Thus, we consider the case that a UAV is used to collect measurements from ground nodes.

A UAV leaves a certain initial location, denoted by $\bu_0$, and, after visiting $N$ nodes, arrives at another location, denoted by $\bu_{N+1}$,
which is referred to as the terminal location. The location of ground node $n$ is denoted by $\bv_n$, $n = 1, \ldots, N$. 
%For convenience, we assume a 2-dimensional space where $\bu_0, \bu_{N+1}, \bv_n \in \uR^2$, $n = 1, \ldots, N$.
The communication region of ground node $n$ is characterized by the following disc:
\be 
\cV_n = \{ \bx: ||\bv_n - \bx||_2 \le r\},
    \label{EQ:cV_n}
\ee 
which is referred to as the communication region of ground node $n$, where $r$ denotes the communication range. 
For each node, there is a possibility of intersecting communication region denoted as 
\begin{equation}
\mathcal{N}(\mathcal{V}_n) = \{\forall j \in N \mid \mathcal{V}_j \cap \mathcal{V}_n \neq \varnothing,~j \ne n \}
    \label{EQ:Intersection}
\end{equation}

It is assumed that the initial and terminal locations do not belong to any of the communication regions of ground nodes, i.e., $\bu_0, \bu_{N+1} \notin \bigcup_n \cV_n$.

%As long as the UAV passes through the communication region of ground node $n$, it is assumed that the UAV can receive a signal from the ground node. % at a location denoted by $\bu_n \in \cV_n$.

The communication range $r$ in \eqref{EQ:cV_n} is determined by the transmit power $P_{\rm g}$ of a ground node. Then, the signal-to-noise ratio (SNR) at a UAV at a distance $r$ becomes
\be  
\sSNR (r) = \frac{P_{\rm g} r^{-\eta}}{N_0},
    \label{EQ:SNR}
\ee 
where $\eta$ and $N_0$ denote the path loss exponent and the noise variance at UAV, respectively. The corresponding achievable rate becomes $C(r) = \log_2 (1+\sSNR(r))$. 

%Note that if we consider a 3-dimensional model, 
For each node, consider a ball with radius $r$, which characterizes the communication region.  In this case, if the altitude of UAV is $h \ (\le r)$, \eqref{EQ:cV_n} can be modified as 
$\cV_n = \{ \bx: ||\bv_n - \bx||_2 \le \bar r\}$,
where $\bar r = \sqrt{r^2 - h^2}$. In order to maximize the area of the communication region, the altitude of UAV, $h$, can be as low as possible, i.e., $h = 0$. In this case, according to \cite{Khuwaja18}, there might be obstacles between the UAV and a ground node, which result in non-line-of-sight (NLoS) paths. Thus, the SNR can be lower  than that in \eqref{EQ:SNR} due to longer propagation paths and small-scale fading. While it would be worth to study the impact of $h$ on the performance, in this paper, we assume that $h$ is greater than a certain non-zero threshold so that the UAV can have a line-of-sight (LoS) to a ground node and the SNR in \eqref{EQ:SNR} is valid. %Thus, hereafter, we assume that $\bar r$

If the number of bits to send to the UAV from a ground node is $B$, the UAV needs to fly within the communication region for a time of $T_{\rm c} = \frac{B}{W \log_2 (1+ \sSNR(r))}$, where $W$ is the bandwidth of the signal transmission from a ground node to the UAV.

\begin{myexample}   \label{EX:1}
Suppose that the speed of a drone (as a UAV) is 60 km/hr, while $W = 22$ MHz and $\sSNR(r) = 6$ dB. If $B = 10^4$ bits, $T_{\rm c}$ becomes $0.196 \times 10^{-3}$ sec or $0.196$ msec. In this case, 
within the communication region, the UAV only needs to move $3.27 \times 10^{-3}$ meters. If $r$ is a few ten meters, 
the moving distance of the UAV for receiving data from a ground node is relatively short, meaning that the UAV does not need to travel deep into the communication region, but only
crosses the boundary. % of the communication region.
\end{myexample}

\begin{figure}[h]
\begin{center}
\includegraphics[width=\figwidth]{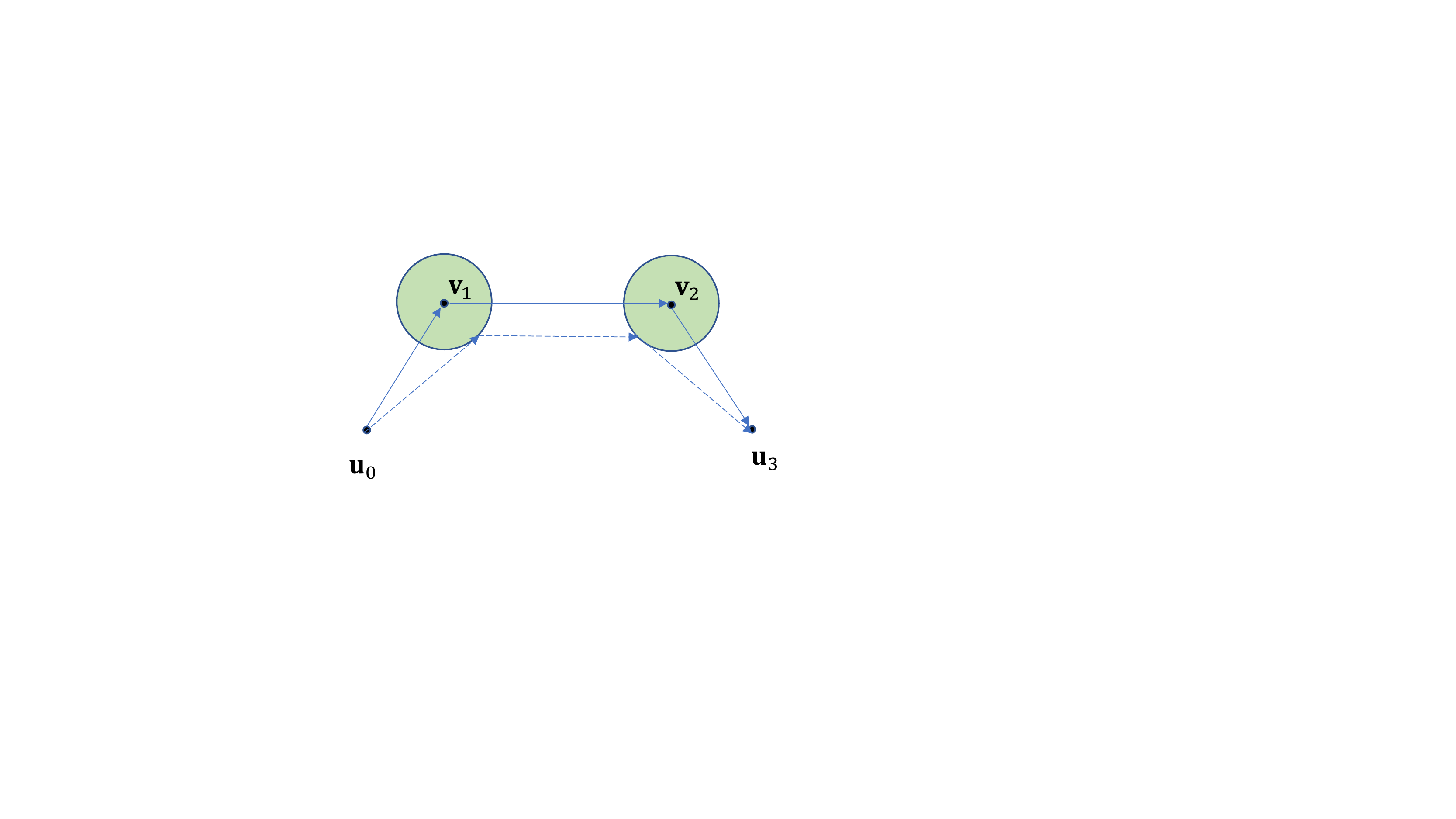} 
\end{center}
\caption{Two different UAV trajectories when $N = 2$.}
        \label{Fig:Ex2}
\end{figure}

\begin{myexample}   \label{EX:2}
In order to see the impact of a non-zero communication range on the travel distance, we consider an example with $N = 2$. As shown in Fig.~\ref{Fig:Ex2}, the trajectory shown by solid lines is the case that the UAV needs to visit the exact locations of two ground nodes, $\bv_1$ and $\bv_2$. On the other hand, the trajectory shown by dashed lines is the case that the UAV just crosses the boundaries of the communication regions of two ground nodes. Clearly, the latter case leads to a shorter travel distance, from which it is expected that the UAV's travel distance decreases with $r$.
\end{myexample}

\section{Problem Formulation}   \label{S:PF}

In this section, we focus on the problem for a UAV that collects data from multiple ground nodes with a non-zero communication range with the following key assumptions:
\begin{itemize}
\item[{\bf A1})] The number of bits, $B$, that a ground node is to send to the UAV is sufficiently small (as illustrated in Example~\ref{EX:1}) so that the data collection time can be ignored when the UAV travels within the communication region.
\item[{\bf A2})] The altitude of the UAV is fixed. As a result, the communication region is characterized by a 2-dimensional circle as shown in \eqref{EQ:cV_n}.
\end{itemize}

For convenience, let $\bu_n \in \cV_n$ be any location that the UAV must pass through to collect measurements or data from ground node $n$, which is referred to as the $n$th collection point. In addition, let $d_{m,n} = ||\bu_{m} - \bu_n||_2$.
There are $N!$ possible paths that the UAV can take. 
For a given path, say path $m \in \{1,\ldots, N!\}$, let $m(n)$ denote the index of the ground node that the UAV visits after visiting $n-1$ ground nodes. Then, the total travel distance of the UAV according to  path $m$ is 
$D_m = d_{0, m(1)} + d_{m(1), m(2)} + \cdots + d_{m(N), N+1}$.
For path $m$, the collection points can be optimized as follows:
\begin{align}
& \{\bu_{m(1)}^\ast , \ldots,\bu_{m(N)}^\ast\}
= \argmin_{\bu_m(n), \ n = 1,\ldots, N} D_m \cr
& \mbox{subject to} \ \bu_{m(n)} \in \cV_{m(n)}, \ n = 1,\ldots, N,
    \label{EQ:inner}
\end{align}
which is referred to as the inner optimization problem. 
Note that due to Assumption of {\bf A1}, the UAV does not need to hover over the communication regions and continues to fly at a constant speed. Thus, if the energy consumption of UAV is proportional to the travel distance, \eqref{EQ:inner} can also be seen as the minimization of UAV energy consumption.

There is also the outer optimization problem that chooses the optimal path with the shortest distance.
As a result, the optimization problem can be given as
\begin{align}
m^\ast & = \argmin_{m \in \{1, \ldots, N!\}} D_m^\ast,
    \label{EQ:all}
\end{align}
where $D_m^\ast$ is the minimum total distance for given path $m$,
which can be obtained by solving \eqref{EQ:inner}.

%Note that if $r = 0$, as mentioned earlier, the overall optimization problem in \eqref{EQ:all} reduces to the conventional TSP, where the inner optimization is not required. Thus, it can be seen that the problem in \eqref{EQ:all} is a generalization of TSP with a non-zero communication range.

\section{Optimization Techniques}

\subsection{Main Results}

In this subsection, we focus on the inner optimization problem and present the main results that can lower the computational complexity.

\begin{mylemma}
For given two points, denoted by $\ba$ and $\bb$, outside of a closed convex set $\cV$, suppose that the straight line between two points does not intersect $\cV$ (if it does, the solution can be readily obtained).
Let
$$
\bz = \argmin_{\bx \in \cV} ||\bx-\ba|| + ||\bx-\bb||.
    \label{EQ:px}
$$
Then, $\bz \in \bar \cV$, where $\bar \cV$ is the boundary of $\cV$.
\end{mylemma} 
\begin{IEEEproof}
Consider the point $\bt$ on the straight line passing through $\ba$ and $\bb$ that satisfies the following:
$
(\bt-\bb)^\rT (\bt-\bz) = 0$.
That is $\bt-\bb$ is orthogonal to $\bt-\bz$. We now assume that $\bz$ is an interior point of $\cV$. By showing that this interior point $\bz$ \emph{cannot} be the solution of \eqref{EQ:px},  we can prove that $\bz$ has to be on the boundary.

%\begin{figure}[h]
%\begin{center}
%\includegraphics[width=\figwidth]{abt.pdf} 
%\end{center}
%\caption{$\bt$ on the straight line passing through two points, $\ba$ and $\bb$, outside a convex set:
%(a) $\bt$ lies on $[\ba, \bb]$, where $[\ba, \bb]$ stands for the straight line between $\ba$ and $\bb$; (b) $\bt$ lies beyond $[\ba, \bb]$.}
%        \label{Fig:abt}
%\end{figure}

By the definition of $\bz$, $||\ba-\bz|| + ||\bb-\bz||$ should be the minimum. Consider another point $\by$ in $\cV$, which is given by
$\by = \bz + \epsilon \bw \in \cV$,
where $\epsilon > 0$ and $\bw = \frac{\bt-\bz}{||\bt-\bz||}$ (thanks to $\cV$ is a closed convex set). Then, there exists $\epsilon$ such that 
$$
||\bt-\by|| = 
||\bt - (\bz + \epsilon \bw)|| = ||(1-\epsilon^\prime) (\bt - \bz)||
<||\bt-\bz||,
$$
where $\epsilon^\prime = \frac{\epsilon}{||\bt-\bz||} < 1$.
From this, by the Pythagorean theorem,
we can also show that
\begin{align*}
||\ba - \bz||^2  & 
= ||\ba - \bt||^2 + ||\bt-\bz||^2 > ||\ba-\bt||^2 + ||\bt- \by||^2 \cr 
& \geq ||\ba - \by||^2 \cr 
||\bb - \bz||^2  & 
= ||\bb - \bt||^2 + ||\bt-\bz||^2 > ||\bb-\bt||^2 + ||\bt- \by||^2 \cr 
& \geq  ||\bb-\by||^2.
\end{align*}
This shows that $||\ba-\bz||+||\bb-\bz|| > ||\ba-\by|| + ||\bb-\by|| $, which contradicts that $\bz$ is the solution. By reductio ad absurdum, this proves that the solution $\bz$ cannot be an interior point, but on the boundary.  
\end{IEEEproof}

\begin{figure}[h]
\begin{center}
\includegraphics[width=\figwidth]{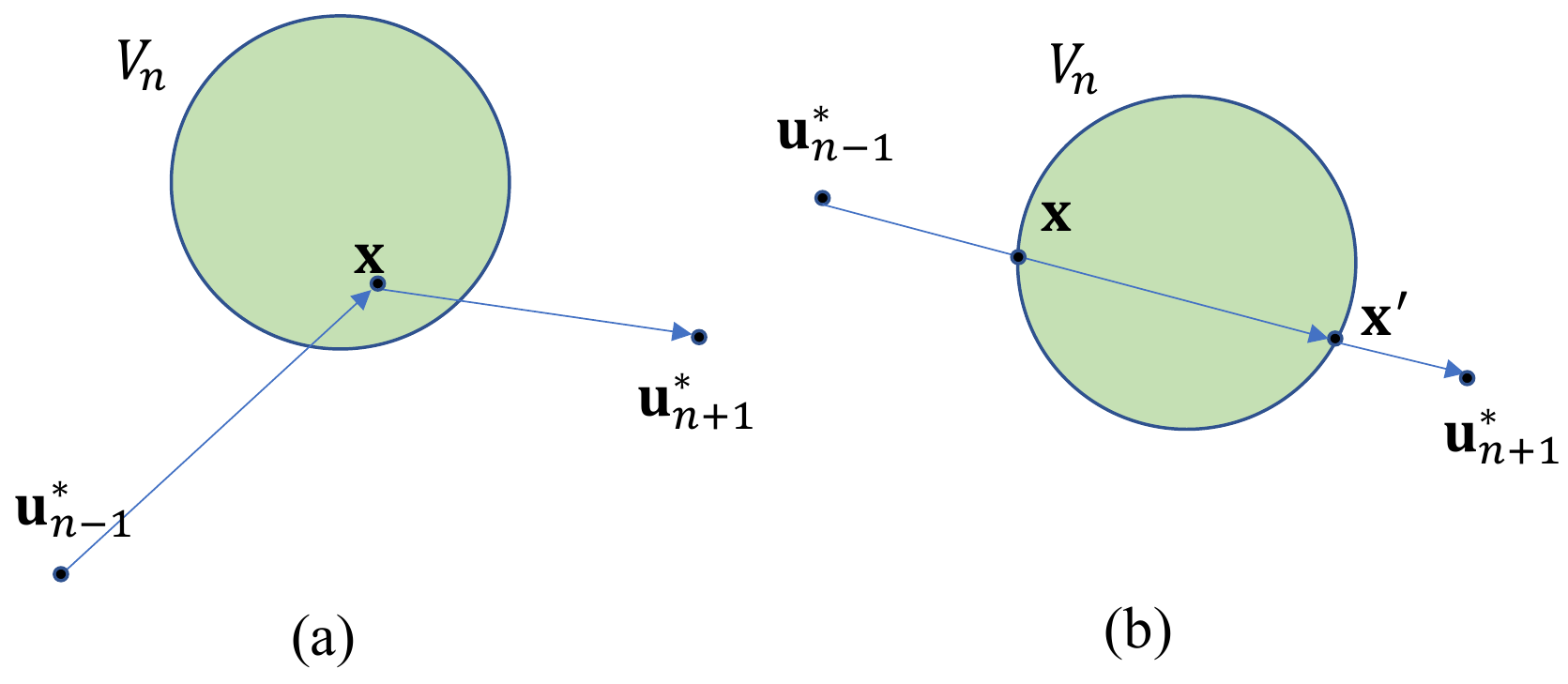} 
\end{center}
\caption{Two possible cases: (a) $\bd \cap \cV_n = \emptyset$; (b) $\bd \cap \cV_n \ne \emptyset$.}
        \label{Fig:P2}
\end{figure}

Let $\bu_{m(n)}^\ast \in \cU_{m(n)}$ denote an optimal collection point of the problem in \eqref{EQ:inner}, where $\cU_{m(n)} \subseteq \uR^2$ is the solution set. 
\begin{mytheorem}   \label{T:1}
If $\cV_n \cap \cV_{n^\prime} = \emptyset$ with $n \ne n^\prime$, for a given path $m$, the optimal collection point $\bu_{m(n)}^\ast$ is not necessarily unique, i.e., $|\cU_{m(n)}|\geq 1$, and there exists at least one optimal collection point that lies on the  circumference of~$\cV_n$.
\end{mytheorem}
\begin{IEEEproof}
For convenience, we assume that $m(n) = n$, $n = 1,\ldots,N$. To show that $\bu_{n}^\ast$ is not necessarily unique, consider $N = 1$ and assume that $\bv_1$ lies on the straight line between $\bu_0$ and $\bu_{N+1} = \bu_2$.  Any point on a straight line passing through the center of the circle $\cV_1$, i.e., $\bv_1$, is an optimal solution if the point is within $\cV_1$. Thus, $\bu_1^\ast$ is not unique.

Based on the induction, we can show that 
there is at least one optimal collection point on the circumference.
Since $\bu_{N+1}$ is a fixed point, we assume that $\bu_{N-1}^\ast$ is also fixed. Once
we show that $\bu_N^\ast$ can be a point on its circumference of $\cV_N$, we can move to $\bu_{N-1}^\ast$. 
That is, for a given (now) $\bu_N^\ast$, assume that $\bu_{N-2}^\ast$ is fixed. Then, it can be shown that $\bu_{N-1}$ has also to be a point on the circumference, and so on. Finally, since $\bu_0$ is a given fixed point, we can show that all $\bu_n^\ast$, $n = 1,\ldots, N$, are points on their circumferences. As a result, what we need to prove is that for given fixed two points, say $\ba = \bu_{n-1}^\ast$ and $\bb = \bu_{n+1}^\ast$, outside of a set $\cV_n$, $\bu_n^\ast$ lies on the boundary of $\cV_n$.
That is, if
\be 
\bu_n^\prime = \argmin_{\bx \in \cV_n} ||\ba - \bx||^2 
+||\bb - \bx||^2 , \label{Eq:NextPoint}
\ee 
we have $\bu_n^\prime \in \bar \cV_n$, where $\bar \cV_n$ is the boundary of $\cV_n$.
There are two possible cases as follows: (a) 
$\bd \cap \cV_n = \emptyset$; and (b) $\bd \cap \cV_n \ne \emptyset$, where
$\bd$ denotes the straight line passing through $\ba$ and $\bb$. Cases (a) and (b) are illustrated in Fig.~\ref{Fig:P2} (a) and (b), respectively. Case (b) is trivial as $\bu_n^\ast$ is any point in the intersection of $\bd$ and $\cV_n$ including the two points, $\bx$ and $\bx^\prime$ on the straight line $\bd$.
For Case (a), we have Lemma 1 to prove. This completes the proof.
\end{IEEEproof}

According to Theorem~\ref{T:1}, the inner optimization problem can be modified as follows.
For path $m$, the collection points can be optimized as follows:
\begin{align}
& \{\bu_{m(1)}^\ast , \ldots,\bu_{m(N)}^\ast\}
= \argmin_{\bu_m(n), \ n = 1,\ldots, N} D_m \cr
& \mbox{subject to} \ \bu_{m(n)} \in \bar \cV_{m(n)}, \ n = 1,\ldots, N,
    \label{EQ:inner_B}
\end{align}
The feasible set of the inner optimization in \eqref{EQ:inner} is $
\cF = \cup_n \cV_n \subseteq \uR^{2N}$,
which can be replaced with
$\cF = \cup_n \bar \cV_n \subseteq \uR^{N}$
when the inner optimization in \eqref{EQ:inner_B} is considered. This results in a decrease of computational complexity when solving the inner optimization problem, while the UAV can successfully perform data collection by crossing the boundaries of communication regions as illustrated in Example~\ref{EX:2}.

\begin{figure}[h]
\begin{center}
\includegraphics[width=\figwidth]{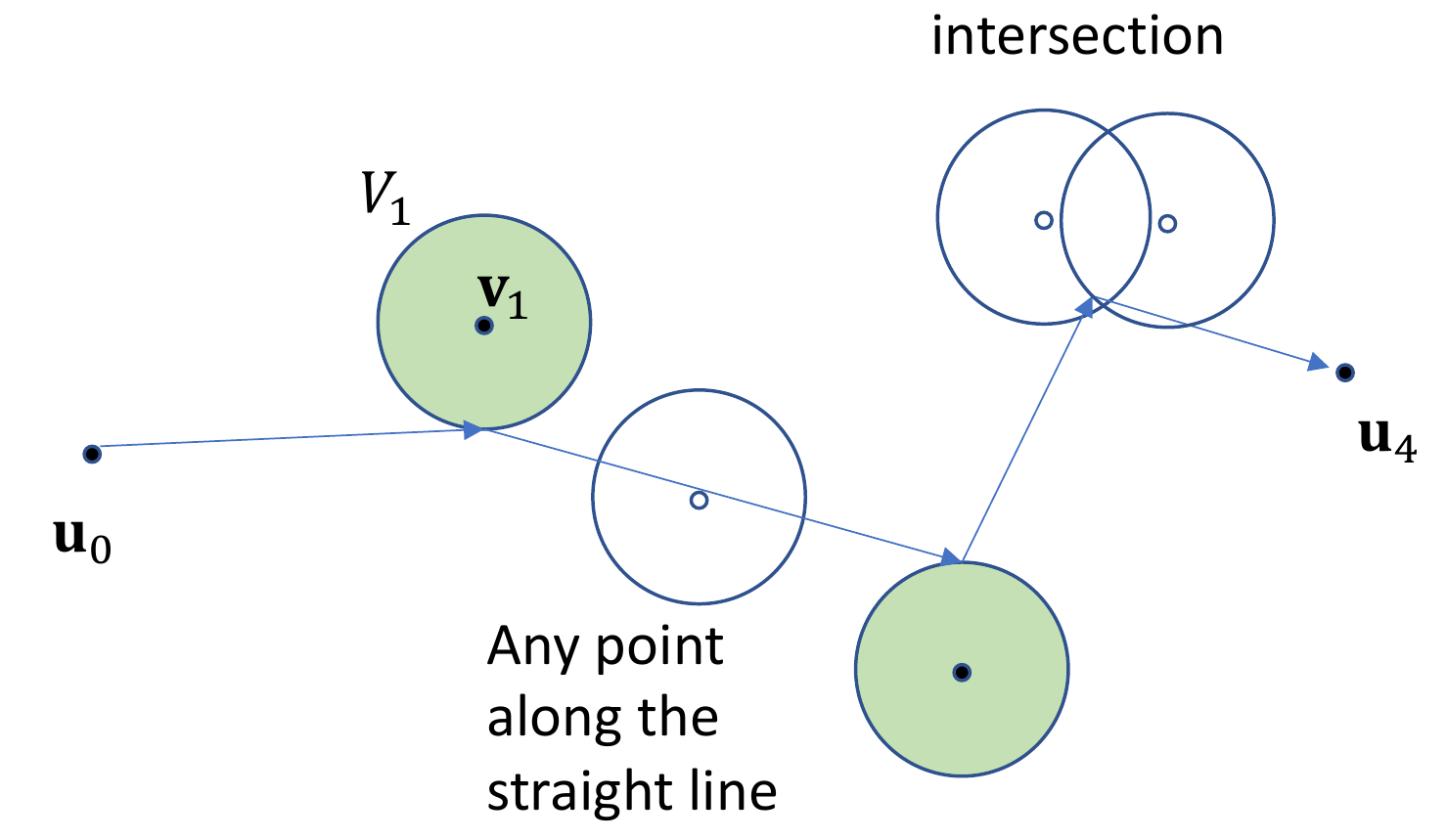} 
\end{center}
\caption{All the cases of communication regions.}
        \label{Fig:cases}
\end{figure}

In Theorem~\ref{T:1}, it is assumed that the communication regions of ground nodes do not overlap. However, if ground nodes are randomly deployed, there can be ground nodes that are closely located so that their communication regions can overlap as illustrated in Fig.~\ref{Fig:cases}. In this case, the UAV can pass the intersection of the ground nodes' communication regions (or cross the boundary of the intersection) to collect data from all of them, because the data collection time is sufficiently short (according to Assumption of {\bf A1}). As a result, we only need to consider the intersection rather than individual communication regions. A generalization of Theorem~\ref{T:1} is given below.

\begin{mytheorem}   \label{T:2}
Suppose that some $\cV_n$'s have nonempty intersection when the corresponding ground users are closely located. The intersection of those communication regions becomes the area that the UAV can collect the data of all the associated ground users, which form a cluster. If $\cV_n$ has no intersection with other $\cV_{n^\prime}$ , $n \ne n^\prime$, itself forms a cluster. Then, there are $\bar N \le N$ clusters, denoted by $\cW_n$, and the UAV needs to pass through $\bar N$ clusters. 
Then, there exists at least
one optimal collection point that lies on the boundary of $\cW_n$.
\end{mytheorem}
\begin{IEEEproof}
The proof is identical to that of Theorem~\ref{T:1} as Lemma 1 is valid as long as the $\cW_n$'s are convex sets. Since the intersection of closed convex sets is also a closed convex set
\cite{Boyd}, any $\cW_n$ is a closed convex set, which completes the proof.
\end{IEEEproof}

\subsection{Numerical Techniques} % to Find Solution}

Based on our findings from the previous subsection,  when searching for optimal solutions, it is sufficient to examine the boundaries of the feasible solution sets.
In this subsection, we discuss numerical techniques to perform the inner optimization for a given order of nodes that the UAV visits. For simplicity, we do not consider \eqref{EQ:all}, but two steps: \emph{i)} an order of sensor nodes is found by solving the conventional TSP using any well-known algorithm
(e.g., for numerical results in Subsection~\ref{SS:Num}, an open source library the Google OR-Tools %\cite{ORToolsG81:online} 
is  used);
\emph{ii)} the optimal data collection point for each sensor or intersection of sensors (that are in close proximity) is determined by solving \eqref{EQ:inner_B}.

%(findingthe conventional TSP and approaches can be used

%The purpose of these experiments is to demonstrate how a two-fold optimization approach can efficiently propose a generalized solution to the TSP for data collection using a UAV. Furthermore, we aim to empirically demonstrate how varying communication ranges can impact the performance of the solution, and to investigate the relationship between communication range and overall cost function reduction. We consider the conventional TSP (r = 0), as a baseline to compare our results. 

%Note that as mentioned earlier, we first need to perform the outer optimization to decide the path or the order of collection as in \eqref{EQ:all}, which is to solve t
%For solving the inner optimization problem \eqref{EQ:inner_B}, we employ a brute-force approach where we look at two scenarios for determining the candidate points

To solve \eqref{EQ:inner_B}, we consider a finite set of $K$ points on each boundary by quantizing them. Thus, a better approximation is obtained with a larger $K$ at the cost of computational complexity.
The details of the implementation are shown in the following pseudocode.

\begin{algorithm}\label{algo:Inner}
\caption{Generating $K$ candidate points on the boundary}
% \\ $\mathcal{Q}$ represents the quantization operation \\
% $\mathcal{I}$ determines the intersecting points between two circles \\
% $\mathcal{A}$ determines the angle of a point from the center of a circle \\
% $\tau$ is the permutation operation to solve equation \eqref{EQ:inner_B}} \Comment{Check intersections and implement proposed algorithm}
\footnotesize\begin{algorithmic}[1]
\Input: $m^\ast, v_{m(n)}, r_{m(n)}, \mathcal{V}_{m(n)}, \mathcal{N}(\mathcal{V}_{m(n)})$
\EndInput
% r_{m(n), \mathcal{V}_{m(n)}$

% \mathcal{N}(\mathcal{V}_{m(n)}), K$
\Output:  $u_{m(n)}^\ast$
% /\quad \quad \quad {Index of sensor nodes based on $m^\ast$}
% \Desc{$v_{m(n)}$} \quad \quad \quad {Location of sensor nodes}
% \Desc{$r_{m(n)}$} \quad \quad \quad {Communication ranges of the sensor nodes}
% \Desc{$\mathcal{V}_{m(n)}$} \quad \quad \quad {Communication area of sensor nodes}
% \Desc{$\mathcal{N}(\mathcal{V}_{m(n)})$} \quad \quad \quad {Intersecting nodes}
% \Desc{$K$} \quad \quad \quad {Candidate access points}

% \Output:
% \Desc{$D_m^\ast$} \quad \quad \quad {Optimal objective function}
% \Desc{$u_{m(n)}^\ast$} \quad \quad \quad {Optimal data collection}
\EndOutput
\Initializations: $u_{m(n)} \gets \emptyset,~p_{m(n)} \gets \emptyset, ~\phi_{m(n)} \gets \emptyset$
\EndInitializations
\State $\mathcal{Q}(v_1, r_1, \theta = [\phi_1, \phi_2], K) \gets$ Quantize equidistant K points
\State $\mathcal{I}(v_1, r_1, v_2, r_2) \gets$ Intersection points between two circles
\State $\mathcal{A}(v_1, p_1) \gets$ Calculate radians from centre and boundary point 
% \ \State $u_{m(n)} \gets \emptyset$     \Comment{Store potential access points}
% \ \State $p_{m(n)} \gets \emptyset$     \Comment{Store intersection points}
% \ \State $\phi_{m(n)} \gets \emptyset$  \Comment{Start/End angle of intersection arc}
% \ \State $\text{int} \gets \emptyset$ \Comment{initialize empty set for intersections}
% \ \State $\text{inter} \gets 0$ 

\For{$n = 1$ to $N$}
\If {$\mathcal{N}(\mathcal{V}_{m(n)}) = \emptyset$} 
\State $u_{m(n)} \gets$ $\mathcal{Q}(v_{m(n)}, r_{m(n)}, \theta = [0, 2\pi], K)$ \\ \Comment Between 0 and 2$\pi$ radians

% set of $k$ equidistant points on the circle \\ \quad \quad \quad boundary using equation (to be linked).

% \State inter = inter + 1
\Else
\For{$i \in \mathcal{N}(\mathcal{V}_{m(n)})$}
\State $p_{m(n)}$ = $p_{m(n)} \cup \mathcal{I}(v_{m(n)}, r_{m(n)}, v_{m(i)}, r_{m(i)})$ \\
\Comment{All intersecting points}
% \mathcal{N}(\mathcal{V}_{m(n)})}$
\EndFor
\For{$p \in p_{m(n)}, i  \in \mathcal{N}(\mathcal{V}_{m(n)})$}
\If{$p$ not in $\mathcal{V}_{m(i)}$} \Comment{Determine intersecting arc}
\State $p_{m(n)} = p_{m(n)} - p$
\EndIf   
\State $\phi_{m(n)} = \phi_{m(n)} \cup \mathcal{A}(p, v_m(n))$
\EndFor 
\State $u_{m(n)} \gets$ $\mathcal{Q}(v_{m(n)}, r_{m(n)}$, $\theta = [\text{min}(\phi_{m(n)})$
\State , $\text{max}(\phi_{m(n)})], K~/\mid \mathcal{N}(\mathcal{V}_{m(n)}) \mid)$ \\ \Comment{Quantize points on the intersecting arc}

% \State $u_{m(n)} \gets$ $\mathcal{Q}(v_{m(n)}, r_{m(n)}, \theta = [min(\phi_{m(n)}), max(\phi_{m(n)})], k/|\mathcal{N(\mathcal{V}_{m(n)})}|)$
\EndIf
\EndFor 
\State $\{u_{m(1)}^\ast, \cdots, u_{m(N)}^\ast\}$ = \textbf{Solve}$(u_{m(1)}, \cdots, u_{m(N)})$ using \eqref{EQ:inner_B} 
\end{algorithmic}
\end{algorithm}

\section{A Case Study: Data Collection for Smart Farming}

In this section, we consider a case study with
$N$ sensor nodes that are distributed over an area of operation (AO) of a river, where the AO is a rectangular area of a length of $L$ (km) and a width of $W$ (km). Each sensor node is able to capture environmental data and transmit it $M$ times a day. This data can help farmers make informed decisions. However, each node is powered by a battery and 
may not be able to send data directly to a remote AP or the BS in order to conserve energy and prolong the battery life.
As a result, as discussed earlier, a UAV as a data mule is used to collect the data from all the nodes and upload it at the BS.

%\begin{figure}[h]
%\begin{center}
%\includegraphics[width=\figwidth]{river.PNG} 
%\end{center}
%\caption{Data collection operation using a UAV for $N$ distributed %sensor nodes.}
%        \label{Fig:cases}
%\end{figure}

\subsection{Energy Consumption Model}

There are two different kinds of energy consumption to take into account as follows.

\subsubsection{UAV} Energy consumption model for an UAV
%, $E_\text{uav}$ is given as 
%    \begin{equation}
%        E_\text{uav} = \alpha(E_\text{comm} + E_\text{idle}) + \beta(E_\text{D}), \beta \gg  \alpha
%        \label{EQ:UAVE}
%    \end{equation}
is mainly based on the amount of energy used by the UAV to travel the necessary distance for collecting data from all nodes and arriving at the BS, which is given by
    \begin{equation}
        E_\text{uav} = cD - c_0,
        \label{EQ:Eu}
    \end{equation}
where $c$ and $c_0$ are constants and $D$ is the travel distance of the UAV \cite{Yan22}.
In this paper, we  only consider a specific UAV, i.e., the DJI Mavic 3 \cite{DJI}. The values of key parameters are given in Table~\ref{table:tab1}, where we can also see that the maximum energy of UAV, denoted by $E_\text{uav}^{\text{max}}$, is limited (due to its battery capacity).

\begin{table}[htb!]
\centering
\caption{Parameter settings for the UAV}
    \begin{tabular}{c|c}
    \hline
    \textbf{Parameter} & \textbf{Values} \\ \hline \hline
         Type & DJI Mavic 3 \cite{DJI}  \\ \hline
         Maximum energy ($E_\text{uav}^{\text{max}}$) & 213, 444 J(oule)  \\ \hline
         Maximum distance ($D^{\text{max}}$) & 30 km \\ \hline
         Constant velocity (V) & 60 km/hr \\
         \hline
\end{tabular}
\label{table:tab1}
\end{table}

\subsubsection{Sensors} The energy consumed by a sensor node with a communication range, $r$, denoted by $E_{\rm s}$, based on \eqref{EQ:SNR}, is given by
\be 
E_{\rm s} = P_{\rm g} T_{\rm c} = r^\eta \overline{\sSNR} N_0 T_{\rm c},
    \label{EQ:Es}
\ee 
where $\overline{\sSNR}$ represents the target SNR. In Table~\ref{table:tab2}, we consider key parameters for a specific sensor node, where the peak transmit power of sensor, denoted by $P_g^{\text{max}}$, is limited.

\begin{table}[htb!]
\centering
\caption{Parameter settings for the UAV}
    \begin{tabular}{c|c}
    \hline
    \textbf{Parameter} & \textbf{Values} \\ \hline \hline
         Type & RN2903 \cite{Lib}  \\ \hline
         Maximum transmission power ($P_g^{\rm max}$)& 18.5 dBm  \\ \hline
         % Maximum communication range ($r^{\text{max}}$) & 15 km \\ \hline
         Protocol & LoRaWAN, Class A \\
         \hline
\end{tabular}
\label{table:tab2}
\end{table}
Consequently, the overall cost function can be defined as a weighted total energy, i.e., $C(r) = \lambda E_{\rm s} + E_{\rm uav}$, where $\lambda \ge 0$ is the weighting factor that determines the relative importance of the two energy terms. Since the cost of replacing batteries at nodes can be expensive, we expect to have a large $\lambda$ (i.e., $\lambda \ge 1$). 

Through Example~\ref{EX:2}, we can see that an increase in $r$ can lead to a decrease in $D$ and a decrease in $E_{\rm uav}$, but as shown in \eqref{EQ:Es}, increasing $r$ may result in a significant increase in $E_{\rm s}$. Thus, we aim to optimize $r$ to  minimize the cost function while taking into account the constraints $E_\text{uav}^{max}$ for the UAV and $P_g^{\rm max}$ for a sensor. Note that due to a finite $P_g^{\rm max}$, we have $r^\eta \overline{\sSNR} N_0 \le P_g^{\rm max}$
or $r \le r_s^\ast$, where $r_s^\ast$ is the maximum communication range with $P_g^{\rm max}$ for given target SNR. In addition, the travel distance of UAV decreases with $r$. 
In particular, if $r = 0$, the travel distance of UAV becomes maximum and the resulting $E_{\rm uav}$ can be greater than the total energy of UAV, $E_\text{uav}^{\text{max}}$. Thus, we have $r \ge r^\ast_{\rm uav}$, where  $r^\ast_{\rm uav}$ represents the minimum communication range to satisfy $E_{\rm uav} \le E_\text{uav}^{\text{max}}$.
Consequently, we  have the following feasible solution of $r$:
\be 
r^\ast_{\rm uav} \le r \le r_s^\ast,  
    \label{EQ:rr}
\ee
where $r^\ast_{\rm uav} \le r_s^\ast$, and the optimal $r$ minimizing $C(r)$  subject to \eqref{EQ:rr} for energy-efficient data collection by UAV can be found as
\begin{equation}
r^\ast = \argmin_{r^\ast_{\rm uav} \le r \le r_s^\ast}    C(r).
    \label{EQ:Cost}
\end{equation}

\subsection{Numerical Results}  \label{SS:Num}

In this subsection, we present numerical results for the case study described in the previous subsection with the AO of $(W, L) = (20, 20)$ in km. %, where $N$ sensors are randomly deployed within the AO.

\begin{figure}[h]
\begin{center}
\includegraphics[width=\figwidth]{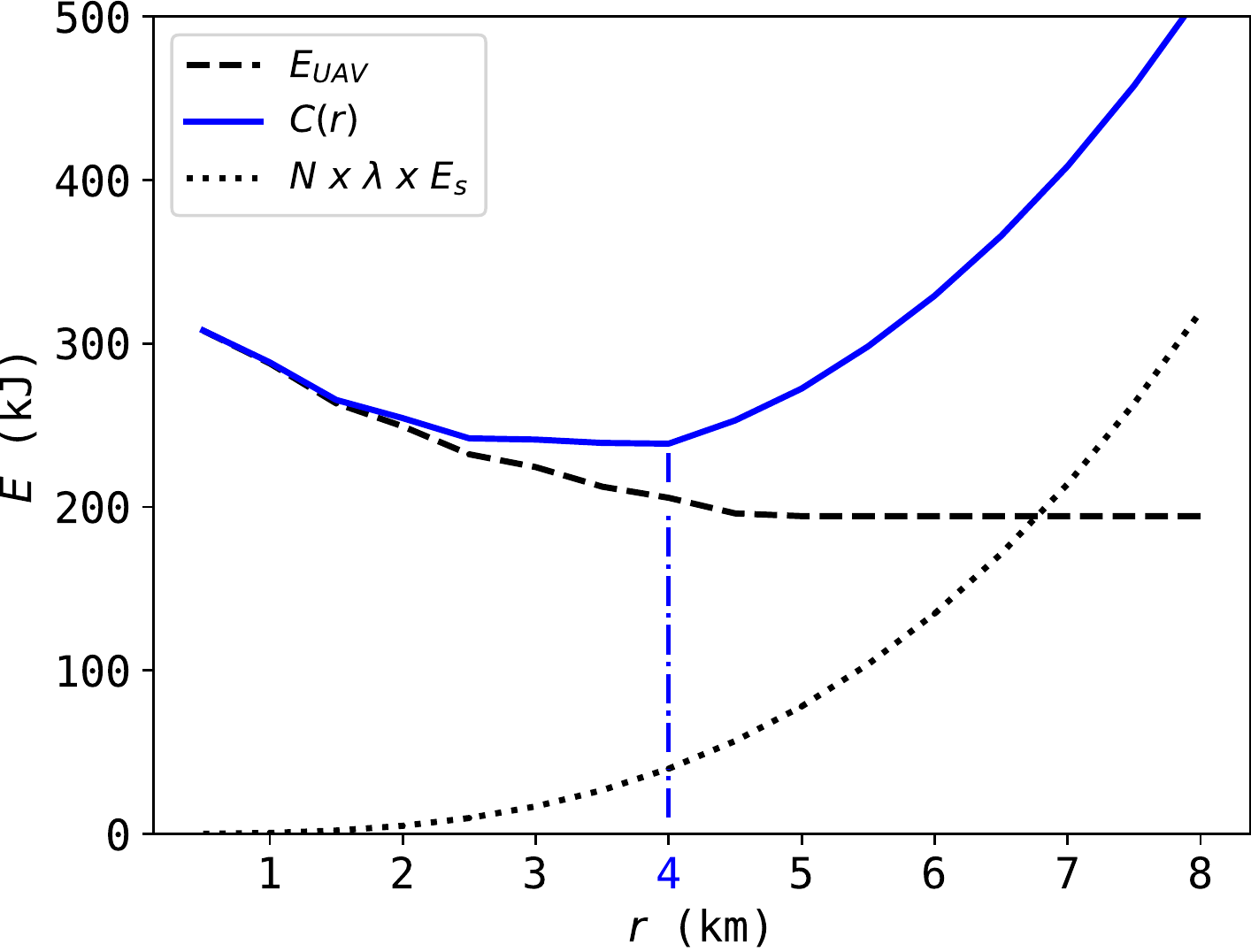} 
\end{center}
\caption{The energy consumed by the UAV and sensors with $\overline{\sSNR} = 6$ dB and  $N = 8$ for different values of communication ranges.}
        \label{Fig:rstar_2}
\end{figure}

Fig.~\ref{Fig:rstar_2} shows the consumed energy curves of the UAV and sensors when $\overline{\sSNR} = 6$ dB and  $N = 8$ as $r$ varies from 0 to 8 km. As expected, the energy consumed by the UAV decreases with $r$, while that by $N$ sensors increases with $r$. As a result, the total weighted energy curve (with $\lambda = 10^{6}$) has a U-shape and an optimal communication range that minimizes the total energy consumption can be found. Note that in this result, the energy consumed by the UAV does not approach 0 although $r$ increases. Since it is assumed that the UAV will fly regardless of $r$, the minimum consumed energy of the UAV cannot be 0, but becomes the value corresponding to the shortest travel distance between $\bu_0$ and $\bu_{N+1}$.

\begin{figure}[!t]
\begin{center} \hspace{-1.5em} \subfigure[$r^\ast$ for varying $N$]{\label{fig 2 ax}\includegraphics[width=0.25\textwidth]{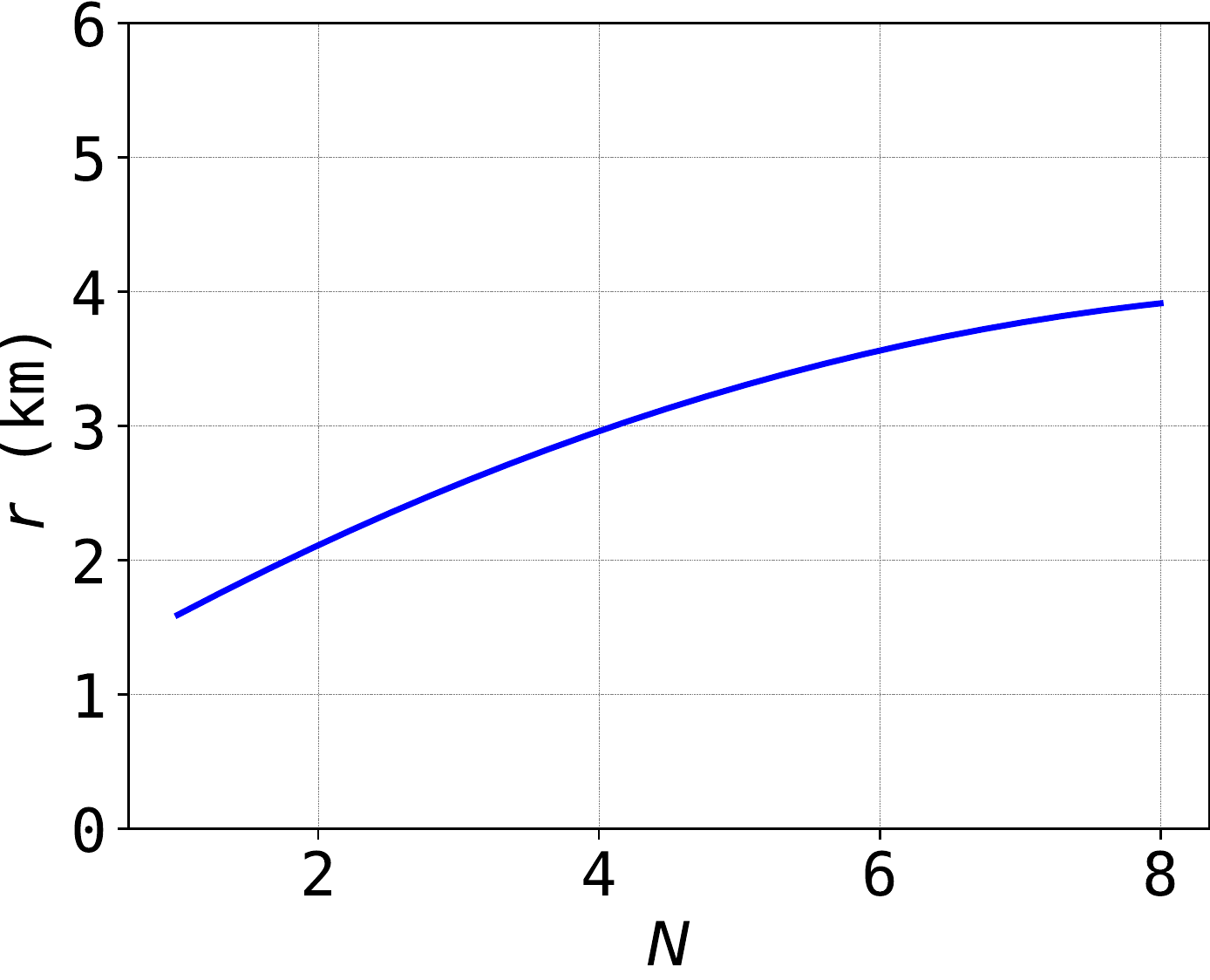}}
\subfigure[$r^\ast$ for varying $\overline{\sSNR}$]{\label{fig 2 bx}\includegraphics[width=0.25
\textwidth]{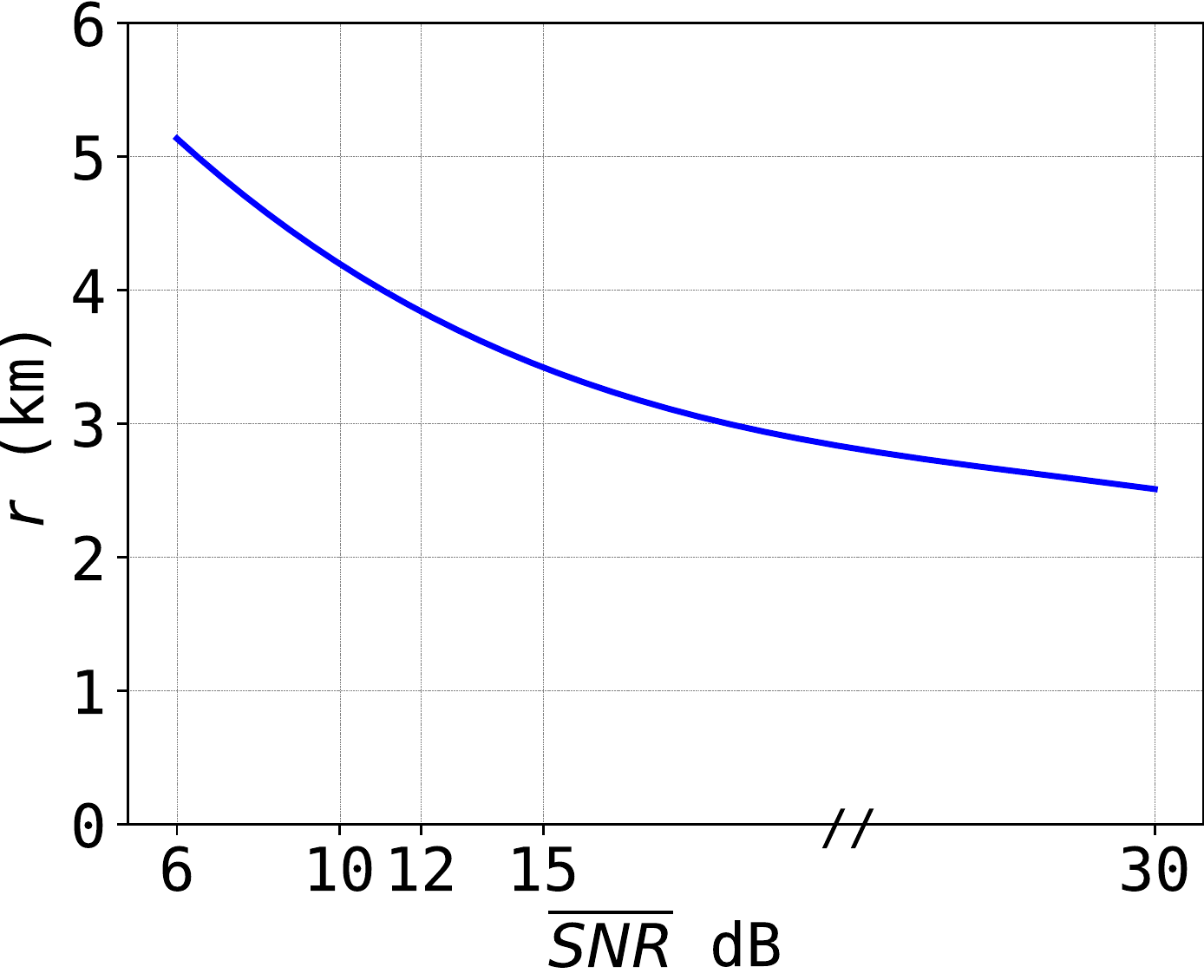}} 
\end{center}
\caption{Optimal communication range of sensors, $r^\ast$: (a) $r^\ast$ as a function of $N$ with $\overline{\sSNR} = 15$ dB; (b) $r^\ast$ as a function of $\overline{\sSNR}$ with $N = 5$.}\label{Fig:ORange}
\end{figure}

Fig.~\ref{Fig:ORange} shows the optimal communication range under different conditions. With $\overline{\sSNR}=6$ dB, $r^\ast$ is shown in Fig.~\ref{fig 2 ax} for different values of $N$. As the number of sensors increases, the energy consumed by the UAV for data collection can increase. To mitigate this increase, $r^\ast$ can increase with $N$. In Fig.~\ref{fig 2 bx}, with $N = 5$, $r^\ast$ is found as as a function of $\overline{\sSNR}$. To meet the required SNR, a shorter $r^\ast$ is expected when $\overline{\sSNR}$ increases.
%There exists a trade-off between the \emph{cost function and the energy consumed by sensor nodes}. As a result, increasing the value of $\overline{\sSNR}$ decreases the optimal communication range. 

\begin{figure}[!h]
\begin{center} 
\includegraphics[width=\figwidth]{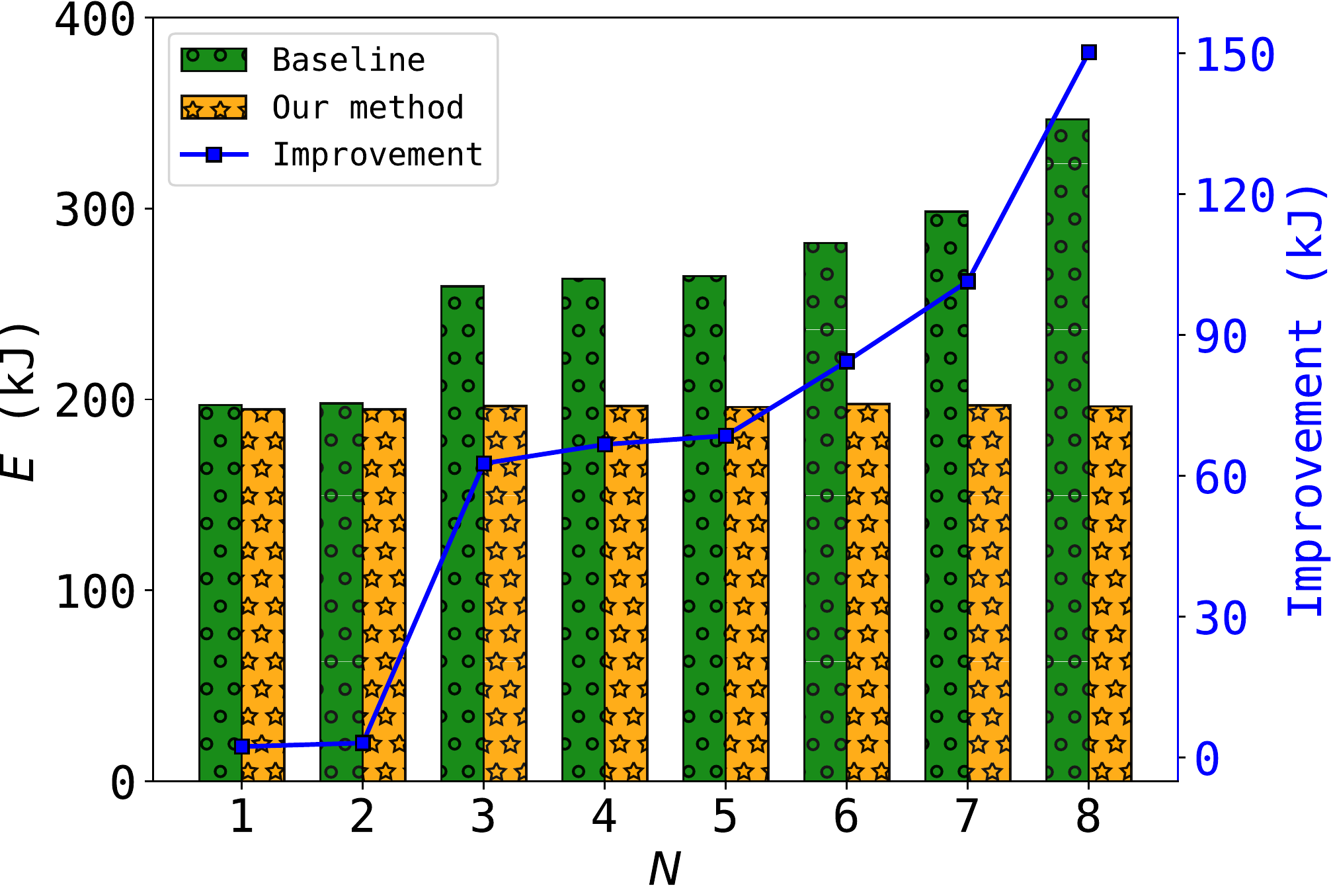} 
\end{center}
\caption{Comparison of energy consumption by UAV using baseline approach versus our method for varying numbers of sensor nodes}
        \label{Fig:rstar_compare}
\end{figure}

In Fig.~\ref{Fig:rstar_compare}, we present a comparison between the energy consumption of a conventional TSP approach and our proposed method for different numbers of sensors. The results indicate that the energy savings achieved by our method become more significant as the number of sensors increases, highlighting the efficiency of our approach, particularly for large values of $N$.  

\begin{figure}[h]
\begin{center} \vspace{2mm}
\includegraphics[width=\figwidth]{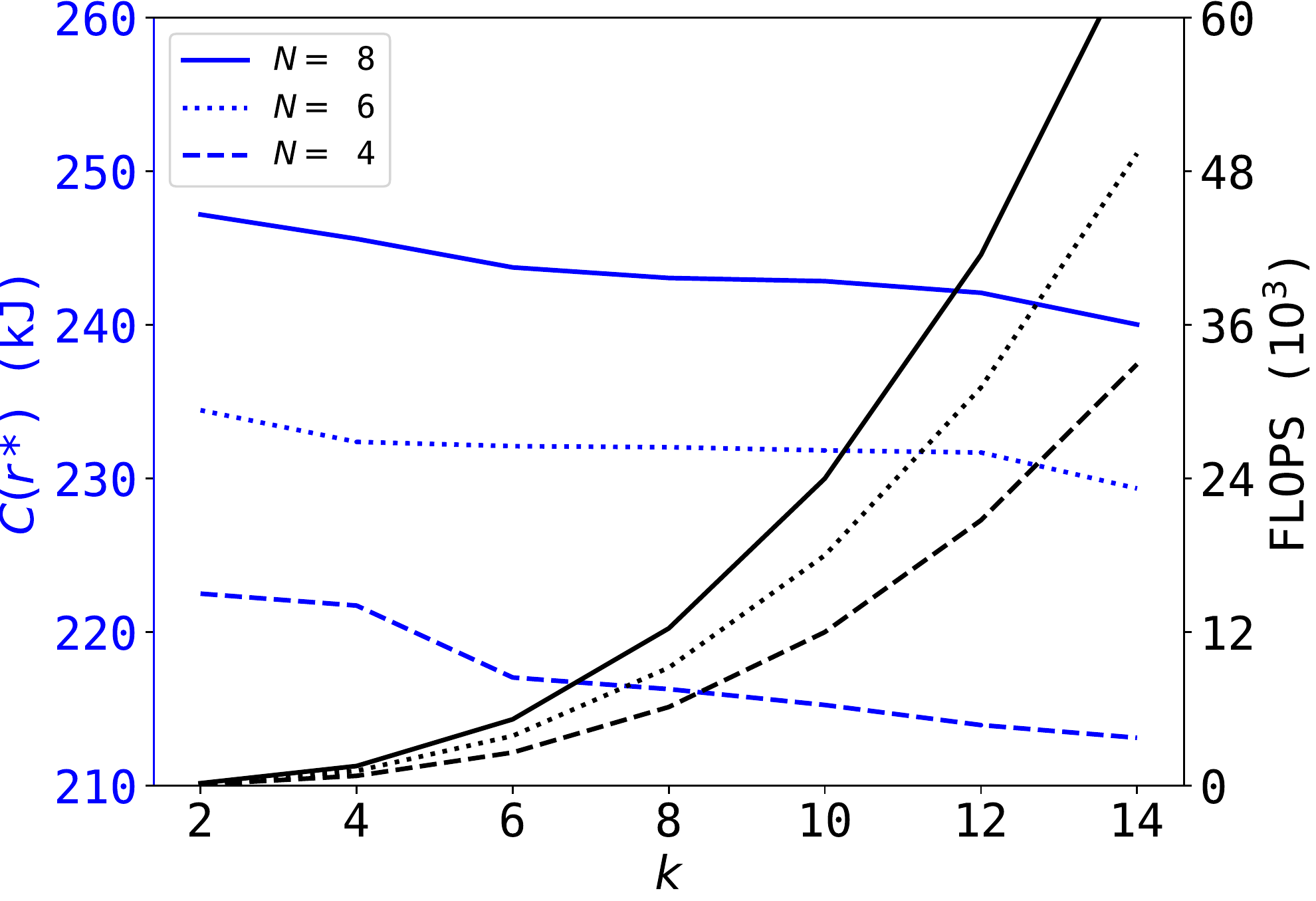} 
\end{center}
\caption{Minimum cost function, $C(r^\ast)$, and computational efficiency (in FLOPS) as a function of candidate points.}
        \label{Fig:FLOP}
\end{figure}

In Fig.~\ref{Fig:FLOP}, we evaluate the impact of the number of candidate points $K$ in minimizing the cost function while also analyzing the computational complexity of our method in terms of Floating Point Operations Per Second (FLOPS).

% \begin{figure}[h]
% \begin{center}
% \includegraphics[width=\figwidth]{rstar_2_3.pdf} 
% \end{center}
% \caption{Optimal communication ranges with N = 2}
%         \label{Fig:rstar_2}
% \end{figure}

% \begin{figure}[h]
% \begin{center}
% \includegraphics[width=\figwidth]{r2_star.png} 
% \end{center}
% \caption{Optimal communication ranges}
%         \label{Fig:rstar_2}
% \end{figure}

\section{Concluding Remarks}

In this paper, we focused on collecting measurements from distributed sensors using a UAV. We rigorously demonstrated that an optimal solution for the generalized TSP can be found at a point on the boundary of each sensor's communication region or at the intersection of communication ranges when sensors are in close proximity. Our key observation suggests that the travel distance decreases with the communication range of sensors. Using this insight, we formulated an optimization problem to minimize the weighted total consumed energy of the UAV and sensors by finding the optimal communication range.

\bibliographystyle{ieeetr}
\bibliography{refs}

\end{document}